\documentclass[10pt,twocolumn,letterpaper]{article}

\usepackage[pagenumbers]{cvpr} 

\usepackage{arydshln}
\usepackage{multirow}
\usepackage{diagbox}

\usepackage{xcolor}         
\usepackage{colortbl}
\definecolor{baselinecolor}{gray}{.9}

\usepackage[most]{tcolorbox}
\definecolor{C1}{HTML}{660874}

\usepackage{makecell}
\usepackage{ulem}
\newcommand{\ours}{CoLoGen}
\definecolor{lightgray}{gray}{.9}
\definecolor{lightblue}{RGB}{230,240,255}
\definecolor{lightgreen}{RGB}{230,255,230}
\definecolor{lightyellow}{RGB}{255,255,230}
\definecolor{lightred}{RGB}{255,230,230}

\definecolor{syxblue}{RGB}{120,130,255}

\usepackage{graphicx}
\usepackage{appendix}
\usepackage{fontawesome}
\usepackage{marvosym}
\newcommand{\envelopeicon}{\textsuperscript{\Letter}}

\definecolor{cvprblue}{rgb}{0.21,0.49,0.74}
\usepackage[pagebackref,breaklinks,colorlinks,allcolors=cvprblue]{hyperref}

\def\paperID{2595} 
\def\confName{CVPR}
\def\confYear{2026}

\title{CoLoGen: Progressive Learning of Concept–Localization Duality for Unified Image Generation}

\author{
    Yuxin Song$^{1}$, Yu Lu$^{2}\envelopeicon$, Haoyuan Sun$^{1,3}$, Huanjin Yao$^{1}$, Fanglong Liu$^{1}$, \\
    Yifan Sun$^{1}$, Haocheng Feng$^{1}$, Hang Zhou$^{1}$, Jingdong Wang$^{1}$\\
    $^{1}$ Baidu Inc. \qquad $^{2}$ Zhejiang University \qquad $^{3}$ Tsinghua University \\
    {\tt\small songyuxin02@baidu.com}
}

\begin{document}
\maketitle

\begin{abstract}
Unified conditional image generation remains difficult because different tasks depend on fundamentally different internal representations. Some require conceptual understanding for semantic synthesis, while others rely on localization cues for spatial precision. Forcing these heterogeneous tasks to share a single representation leads to concept-localization representational conflict.
To address this issue, we propose CoLoGen, a unified diffusion framework that progressively learns and reconciles this concept-localization duality. CoLoGen uses a staged curriculum that first builds core conceptual and localization abilities, then adapts them to diverse visual conditions, and finally refines their synergy for complex instruction-driven tasks. Central to this process is the Progressive Representation Weaving (PRW) module, which dynamically routes features to specialized experts and stably integrates their outputs across stages.
Experiments on editing, controllable generation, and customized generation show that CoLoGen achieves competitive or superior performance, offering a principled representational perspective for unified image generation.
\end{abstract}    
\vspace{-5mm}
\section{Introduction}
\label{sec:intro}

\begin{figure}[t]
\centering
\includegraphics[width=1.0\linewidth]{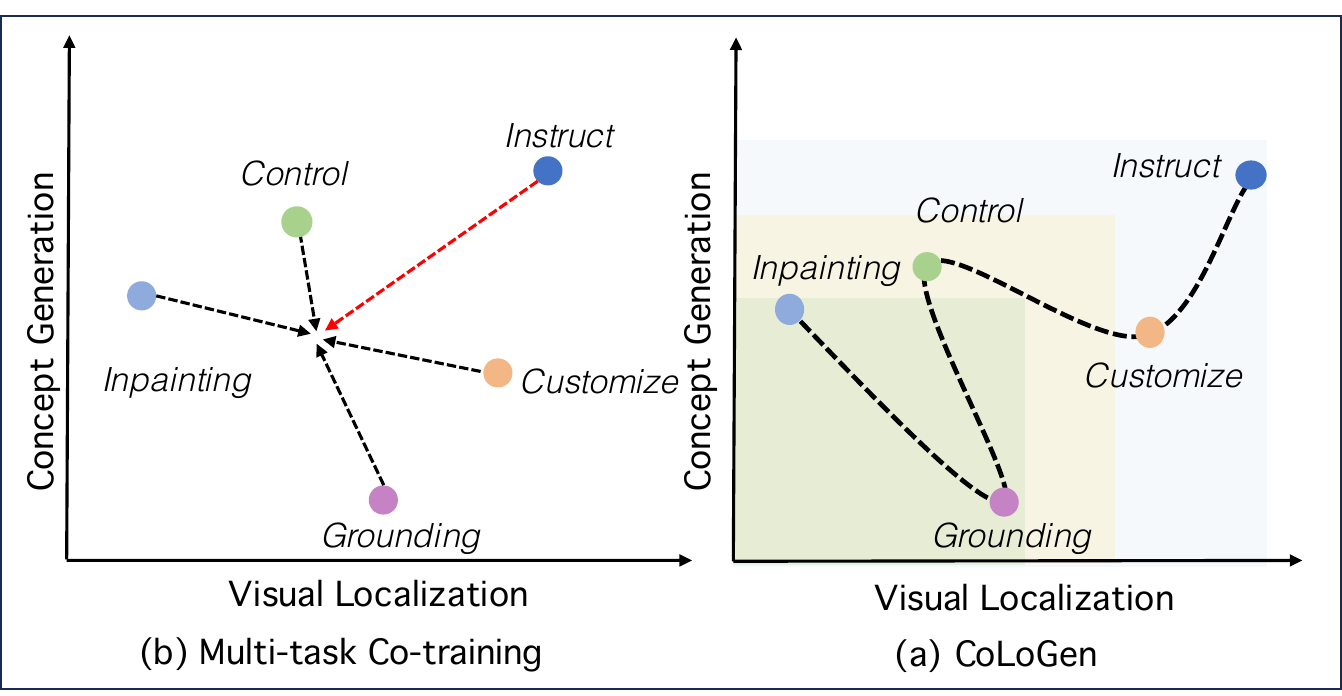}
\vspace{-5mm}
\caption{The comparison between the multi-task learning strategy (a) and the ours progressive staged training (b) within the framework of unified multi-modal image generation. We specifically examines five conventional tasks: mask inpainting, image grounding, controllable image generation, customized image generation, and instruction-based image editing.}
\vspace{-5mm}
\label{fig:intro}
\end{figure}


Unified multimodal image generation~\cite{xiao2024omnigen,lin2024pixwizard,chen2024unireal, OMNIEDIT, liu2024lumina, pan2023kosmos-g, xia2024dreamomni, wu2025qwenimage, zhao2024monoformer, cai2025zimage} has recently attracted significant attention, as models aim to address diverse tasks (such as mask inpainting, image grounding, controllable synthesis, customized generation, and instruction-based editing) within a unified framework. Drawing from the success of unified architectures in language modeling~\cite{gpt3,llama1, GPT-4o-mini, openai2024gpt4o}, recent efforts have sought to develop generalist diffusion models capable of handling varied visual conditions through shared encoders, backbones, or in-context interfaces.

Unlike NLP, where tasks share relatively uniform token-level representations, unified image generation faces a core challenge in representation.
Tasks such as inpainting~\cite{ju2024brushnet,chen2024mimicbrush,painter} or subject-driven generation~\cite{mou2024t2i, ye2023ip, Flux-IPadapter, pan2023kosmos-g} rely heavily on \textbf{conceptual representations}, which encode semantic coherence and object-level understanding.
Conversely, grounding and controllable generation~\cite{controlnet, qin2023unicontrol, zhang2023controlnet} demand \textbf{localization representations} that emphasize spatial alignment, geometry, and structural consistency.
More complex tasks such as instruction-based editing \cite{zhang2024magicbrush, sheynin2024emuedit, OMNIEDIT, xie2024show, xia2024dreamomni, zhang2025incontext, instructblip, brooks2023instructpix2pix} rely on a synergistic integration of both.

We refer to this fundamental conflict as the Concept–Localization Duality:
Conceptual and localization cues occupy competing subspaces in the generative latent space, and naïvely optimizing them jointly leads to representational interference and unstable training~\cite{team2405chameleon,chen2023octavius}.
Such interference explains why existing unified frameworks tend to excel at a subset of tasks while underperforming on others indicating that resolving this Duality is essential for reliable generalist image generation.

Hence, this leads us to the following insight:
\begin{tcolorbox}[enhanced,colback=white,%
    colframe=blue, attach boxed title to top right={yshift=-\tcboxedtitleheight/2, xshift=-.25cm}, title=\large{\textbf{Insight}}, coltitle=blue, boxed title style={size=small,colback=white,opacityback=1, opacityframe=0}, size=title, enlarge top initially by=-\tcboxedtitleheight/2]
\vspace{0.25em}
\textcolor{blue!}{\textbf{\textit{Can we utilize a progressive, easy-to-hard training curriculum to overcome these conflicts?}}
}
\end{tcolorbox}

Based on this, we propose \underline{\textbf{Co}}ncept–\underline{\textbf{Lo}}calization \underline{\textbf{Gen}} eration (\textbf{CoLoGen}), that progressivelly unifies conditional image generation by explicitly structuring all tasks around conceptual and localization representations. As illustrated in Fig.~\ref{fig:intro}(b), CoLoGen employs a \textbf{progressive staged training strategy} to reduce representational conflict. To begin with, it learns fundamental conceptual and localization abilities from large-scale synthetic data through tasks such as mask inpainting and visual grounding \cite{refcoco+, mao2016refcocog, refcoco, zhao2024octopus}. Secondly, it adapts these abilities to diverse conditional signals, such as segmentation, depth, and Canny edges. Finally, it refines their synergy through instruction-image alignment on complex editing and customization tasks. Such a progressive, easy-to-hard curriculum effectively could mitigate the issues of conflicting and significantly improves performance on complex tasks.

Furthermore, to stabilize optimization across stages and preserve acquired knowledge, we introduce a lightweight architectural component termed \underline{\textbf{P}}rogressive \underline{\textbf{R}}epresentation \underline{\textbf{W}}eaving (\textbf{PRW}). 
While recent in-context learning frameworks \cite{chen2024unireal, zhang2025incontext, zhang2025icedit} have unified diverse instructions at the input modality level by concatenating reference images, control signals, and noisy latents, they do not explicitly address underlying representational conflicts during joint training.
PRW resolves representational conflicts by using a pool of lightweight experts that separately acquire concept and localization skills in early training.
A dynamic router, guided by Veteran Gate Routing, then learns how to activate and combine these experts for different tasks.
Through this staged integration, PRW gradually weaves the dual representations into a stable unified space while avoiding catastrophic forgetting.

In summary, this work contributes the following:
\begin{itemize}
\item We propose Concept–Localization Generation (CoLoGen), a unified multimodal image generation framework that alleviates task conflicts through explicit structuring around conceptual and localization representations.
\item We propose a progressive staged learning strategy and a novel Progressive Representation Weaving (PRW) architecture that dynamically routes and integrates specialized experts across training stages.
\item Extensive experiments on instruction editing, subject-driven generation, and controllable image generation demonstrate that CoLoGen achieves performance competitive with or surpassing task-specific state-of-the-art methods.
\end{itemize}

\section{CoLoGen}
\begin{figure*}[ht]
\centering
\includegraphics[width=0.98\linewidth]{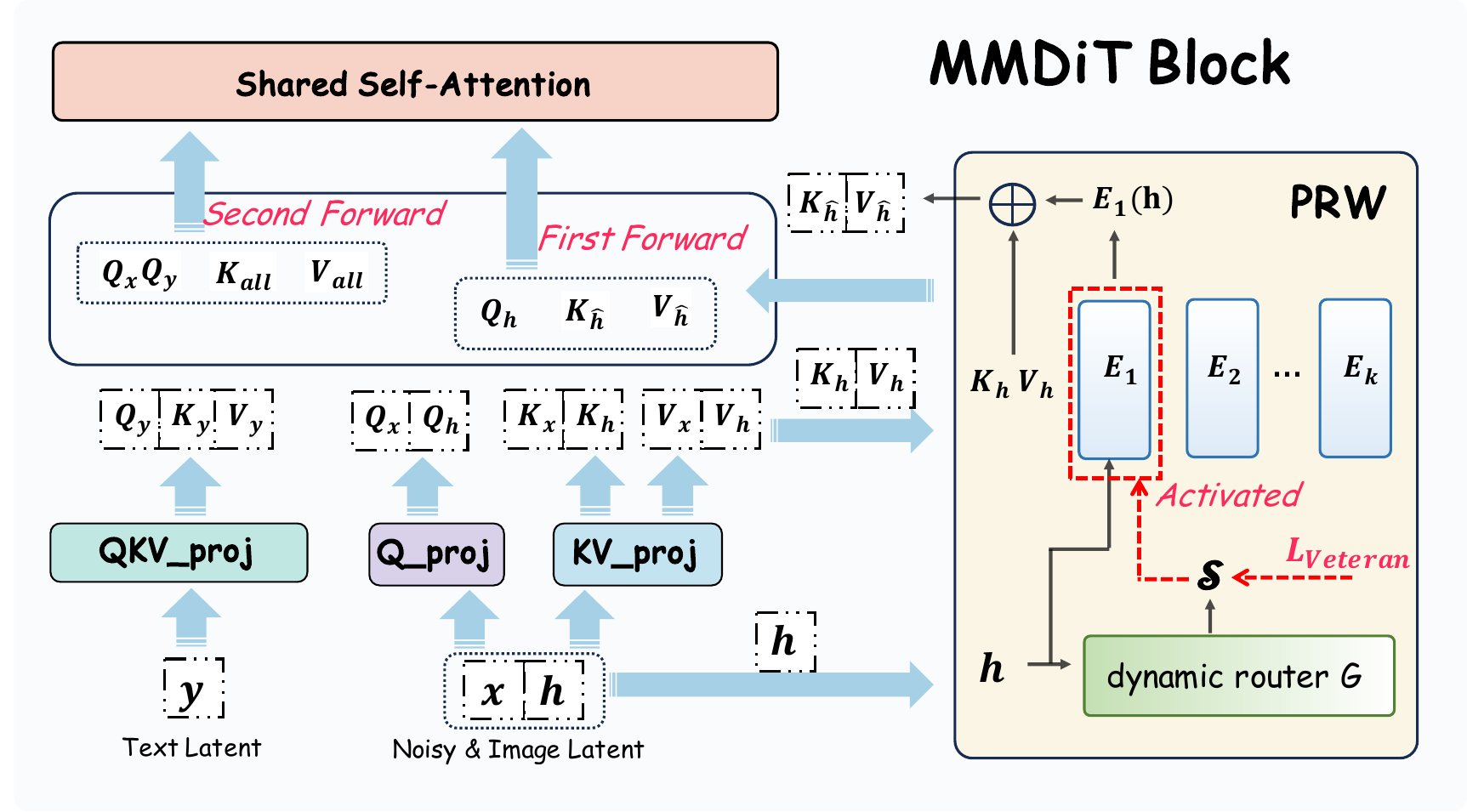}
\vspace{-1em}
\caption{The overall framework of the unified Multi-modal to Image generation model, CoLoGen. For each training stage, CoLoGen efficiently integrates a set of condition-specific experts via the Progressive Representation Weaving (PRW) which are constructed on KV projection layers and a dynamic router $G$. Notably, The QKV projection layer and the self-attention layer are sharing weights for the inputs of Noisy Latent and Source Image Latent. CoLoGen employs a progressive staged training strategy to gradually increase the number of experts $E_k$, allowing it to better adapt to more complex downstream tasks.}
\vspace{-1em}
\label{fig:framework}
\end{figure*}

\subsection{Concept and Localization Representations}
\label{sec:representations}
Conditional image generation tasks fundamentally rely on two complementary abilities: \textbf{visual concept generation} and \textbf{visual localization}. Their relative importance varies by task. 
Specifically, \textit{controllable generation} tasks provide strong structural conditions (e.g., segmentation, edges, or depth), enabling the model to largely disregard localization requirements and instead focus on regenerating coherent visual concepts. 
In contrast, \textit{customized generation} tasks demand precise localization of the subject in a reference image to preserve identity-specific features while allowing a degree of conceptual generalization in the generated output. 
\textit{Instruction-based editing} tasks necessitate not only the comprehension of textual instructions but also accurate localization of the target editing region, followed by the regeneration of visual concepts within that localized area.


\noindent\textbf{Hypothesis.} We formalize these two abilities as two distinct and underlying representations. Let $h \in \mathbb{R}^{L \times d}$ be an intermediate feature map within the diffusion model's transformer blocks. We posit the existence of a \textbf{concept representation} $\mathcal{R}_c$ and a \textbf{localization representation} $\mathcal{R}_l$, which can be extracted from $h$ via the mapping functions $f_c$ and $f_l$, respectively:
\begin{equation}
    \mathcal{R}_c = f_c(h), \quad \mathcal{R}_l = f_l(h)
\end{equation}
Our central hypothesis is that the failure of existing unified models stems from forcing a single, static fusion of these representations across all tasks. This joint optimization creates a representational conflict, where improving the model's capacity for conceptual understanding (optimizing $f_c$) can degrade its spatial precision (harming $f_l$), and vice-versa. A successful unified model must therefore be able to \textit{dynamically} modulate the influence of $\mathcal{R}_c$ and $\mathcal{R}_l$ based on the specific demands of each task.
Despite recent advances, generalist image generation models \cite{chen2024unireal, xiao2024omnigen, zhao2024ultraedit, lin2024pixwizard, chen2023photoverse, shen2024key} have not yet systematically explored the synergistic interaction between visual concept and localization representations. We argue that such a unified representational perspective is critical for advancing the scope and reliability of unified conditional image generation.

\subsection{Model Design}
\label{sec:approach}

To validate this hypothesis, CoLoGen is grounded in two key components: (1) the Progressive Representation Weaving (PRW) architecture, which forms the structural basis for dynamic representation management; and (2) a Progressive Staged Training strategy, which provides the methodological framework to resolve representational conflicts in an easy-to-hard manner. Our overall framework is built upon the advanced FLUX.1 architecture \cite{fluxfill}.

\noindent\textbf{Progressive Representation Weaving (PRW). }To enable dynamic management and integration of conceptual and localization representations, we propose the Progressive Representation Weaving (PRW) architecture. As illustrated in Figure~\ref{fig:framework}, PRW operates within each multi-modal attention block to adapt the source latent $h$ for diverse task demands, complementing the standard processing of the noisy latent $x$ and the text latent $y$. The architecture comprises a dynamic routing mechanism $G$ and a pool of $N$ specialized, parameter-efficient experts $\{E_k\}_{k=1}^N$, where each expert serves as a Key–Value projection module, denoted as $\text{KV\_proj}_{k}$. The router $G$, implemented as a Noisy Router, determines the most suitable expert by generating a vector of pre-softmax logits $\mathbf{w}$ over the expert pool conditioned on the input latent $h$. Formally, this can be defined as:
\begin{equation}
    \mathbf{w} = h W_r + \epsilon \odot \text{softplus}(h W_n), \quad \epsilon \sim \mathcal{N}(0, \mathbf{I}),
\end{equation}
where $W_r$ and $W_n$ are learnable projection matrices for the routing logits and the noise scale, respectively. The term $\epsilon$ represents standard Gaussian noise, and $\odot$ denotes element-wise multiplication. This noise injection during training encourages balanced expert utilization, while the mechanism remains fully deterministic during inference.

Following the logit computation, a sparse activation function is subsequently applied. We apply a top-1 selection on the softmax-normalized logits to identify the single most relevant expert for the given input:
\begin{equation}
    \mathcal{S} = \text{TopK}(\text{Softmax}(\mathbf{w}), n=1),
    \label{eq:s_index}
\end{equation}
where $\mathcal{S}$ is the set containing the index of the activated expert. The resulting adaptive residual is computed by passing the original latent $h$ through the selected expert and scaling the output by its corresponding softmax weight. This residual is then added to a base projection to obtain the final Key and Value representations for the source latent, $K_{\hat{h}}$ and $V_{\hat{h}}$:
\begin{equation}
    (K_{\hat{h}}, V_{\hat{h}}) = \text{KV\_proj}_{\text{base}}(h) + \sum_{k \in \mathcal{S}} \text{softmax}(\mathbf{w})_k E_k(h)
\end{equation}
While the PRW module dynamically generates the Key and Value representations for the source latent, its Query $Q_h$, along with the QKV projections for all other latents, is produced through standard linear projection layers.

\begin{table*}[htbp]
\centering
\renewcommand{\arraystretch}{1.0}
  \setlength\tabcolsep{2mm}
\caption{\textbf{The outline of training data about each training stage.} Notably, Endogenous Pre-training comprises two distinct training steps: mask inpainting and image grounding. Instruction-Image Alignment encompasses both Customized Generation and Instruction Editing. }
\label{tab:recipe}
\centering
\scalebox{1.0}{
\begin{tabular}{l|lll}
\Xhline{1.2pt}
\textbf{Stage} & \textbf{Task} & \textbf{\# Samples} & \textbf{Data Source} \\

\hline
\multirow{2}*{\textbf{Endogenous Pre-training}} & Mask Inpainting  & 3M & ADE20k, COCOStuff, JouneyDB \\
  & Image Grounding & 1M & RefCOCO, RefCOCOg, RefCOCO+, LVIS \\
\hline
\multirow{2}*{\textbf{Conditional Injection}} & \multirow{2}*{Controllable Generation} & \multirow{2}*{20M}  & Multigen-20M, Multigen-20M-Depth, \\
 &  &  & ADE20k, COCOStuff \\

\hline
\multirow{2}*{\textbf{Instruction-image alignment}} & Customized Generation  & 200K  & Subject200k \\
 & Instruction Editing  & 1.6M  & OmniEdit, Magicbrush, \textcolor{gray}{In-house Data} \\
\Xhline{1.2pt}
\end{tabular}
}
\vspace{-2mm}
\end{table*}

The attention mechanism then proceeds in a structured and sequential manner to fuse these representations. First, the source latent's Query ($Q_h$) attends to its own dynamically adapted Key-Value pair $(K_{\hat{h}}, V_{\hat{h}})$ in a self-attention step. This allows the source representation to internalize the task-specific information introduced by the activated expert. Subsequently, the stem self-attention mechanism is employed to update the noisy and text latents. The queries $Q_x$ and $Q_y$ attend to a comprehensive context formed by concatenating the Keys and Values from all three modalities, including the newly adapted source representations:
\begin{align}
    K_{\text{all}} &= \text{concat}(K_x, K_y, {K}_h), \\
    V_{\text{all}} &= \text{concat}(V_x, V_y, {V}_h), \\
    \text{Output}_{x,y} &= \text{Self-attn}(\text{concat}(Q_x, Q_y), K_{\text{all}}, V_{\text{all}}).
\end{align}
This two-stage process ensures that both text and noisy latents can draw upon a source representation that has already been refined for the specific task, enabling a more effective and context-aware fusion of multimodal information.

\noindent\textbf{Progressive Staged Training.} The CoLoGen framework employs a progressive staged training strategy, as illustrated in Figure~\ref{fig:intro}(a), to systematically mitigate representational conflicts and enhance cross-task complementarity. Inspired by principles of lifelong learning, this strategy organizes all tasks around complementary conceptual and localization representations. During this multi-step training, the model progressively develops and strengthens its capabilities.

At each training step $t \in [0,4]$, the model integrates $N$ specialized, parameter-efficient experts $\{E_k\}_{k=0}^{N-1}$ through the Progressive Representation Weaving (PRW) architecture, where $N = t + 1$. For notational simplicity, the subscript $t$ is omitted in the subsequent formulations. The expert $E_{N-1}$ is designated for the task-specific training at step $t$, while other experts remain frozen.

\noindent\textbf{Veteran Gate Routing Supervision. }To effectively leverage knowledge acquired in preceding training steps and encourage balanced expert utilization, we propose a {Veteran Gate Routing Supervision} mechanism. This mechanism incorporates an auxiliary supervision term into the overall training loss, guiding the dynamic routing module to align its expert assignment distributions with desired usage ratios.

Given the defined set of experts $\{E_k\}_{k=0}^{N-1}$ ($N=t+1$), a regularization term is applied to encourage routing according to predefined ratio-specific experts across all MMDiT blocks. As indicated by the sparse activation in Equation \ref{eq:s_index} from the routing module, $\mathcal{S}$ denotes the set of selected expert indices, with $|\mathcal{S}|=1$ in our implementation. The usage ratio of the specific expert $E_{N-1}$ is calculated as:
\begin{equation}
    U_t = \frac{1}{L_n}\sum_{i=1}^{L_n}\mathbb{I}(e_i = N-1),
\end{equation}
where $L_n$ represents the total number of MMDiT blocks, and $e_i \in \mathcal{S}$ indicates the assigned expert for block $i$.
We then define the veteran gate routing supervision loss term $\mathcal{L}_{\text{veteran}}$ to penalize deviations from the desired routing density $\rho$ of specific experts:
\begin{equation}
    \mathcal{L}_{\text{veteran}} = \alpha \cdot |U_t - \rho|,
\end{equation}
where $\alpha$ is a hyperparameter that balances the veteran gate routing supervision loss with the primary diffusion loss.
Consequently, the total training loss $\mathcal{L}_{\text{total}}$ comprises two parts: the primary diffusion generation loss $\mathcal{L}_{\text{task}}$ and the veteran gate routing supervision loss term $\mathcal{L}_{\text{veteran}}$:
\begin{equation}
\mathcal{L}_{\text{total}} = \mathcal{L}_{\text{task}} + \mathcal{L}_{\text{veteran}}.
\end{equation}
This auxiliary supervision, $\mathcal{L}_\text{veteran}$, plays a crucial role in guiding the gating network to preferentially select specific experts, thereby enabling dynamically balanced expert utilization and contributing to a more stable training process.

\subsection{Unified Multi-modal to Image Generation Dataset}

\begin{table*}[htbp]
\centering
\renewcommand{\arraystretch}{1.0}
  \setlength\tabcolsep{2mm}
\caption{\textbf{Quantitative results for instruction image editing} evaluated on the Emu Edit test split and MagicBrush test split. ``-'' indicates that the method does not report the corresponding results. $\!\uparrow$ indicates higher result
is better, while $\!\downarrow$ means lower is better.} 
\label{tab:editing}
\centering
\begin{tabular}{l|cccc|cccc}
\Xhline{1.2pt}
\multirow{2}*{\textbf{Methods}} & \multicolumn{4}{c|}{\bf Emu Edit test set} & \multicolumn{4}{c}{\bf MagicBrush test set}\\
 &  $\text{CLIP}_{i}\!\uparrow$ & $\text{CLIP}_{out}\!\uparrow$ &  $\ell_1\!\downarrow$  & DINO$\uparrow$ &  $\text{CLIP}_{i}\!\uparrow$ & $\text{CLIP}_{out}\!\uparrow$ &  $\ell_1\!\downarrow$ & DINO$\uparrow$ \\  
\hline
\rowcolor{gray!20}
\multicolumn{1}{l|}{\textsl{Specialist Models} } & \multicolumn{4}{l|}{} & \multicolumn{4}{l}{} \\ 
InstructPix2Pix~\cite{brooks2023instructpix2pix} & 0.834 & 0.219 & 0.121 & 0.762 
& 0.837 & 0.245 & 0.093 & 0.767\\
MagicBrush~\cite{zhang2024magicbrush} & 0.838 & 0.222 & 0.100 & 0.776
& 0.883 & 0.261 & 0.058 & 0.871 \\
PnP~\cite{tumanyan2023plug} & 0.521 & 0.089 & 0.304 & 0.153
& 0.568 & 0.101 & 0.289 & 0.220 \\
Null-Text Inv.~\cite{mokady2023null} & 0.761 & 0.236 & 0.075 & 0.678
& 0.752 & 0.263  & 0.077 & 0.664 \\
Emu Edit~\cite{sheynin2024emuedit} & {0.859} & 0.231 & 0.094 & {0.819}
& {0.897} & 0.261 & \textbf{0.052} & \textbf{0.879} \\
\hline
\rowcolor{gray!20}
\multicolumn{1}{l|}{\textsl{Generalist Models}}  & \multicolumn{4}{l|}{} & \multicolumn{4}{l}{} \\ 
UltraEdit~\cite{zhao2024ultraedit} & 0.844 & 0.283 & 0.071 & 0.793
& 0.868 & -  & 0.088 & 0.792 \\
OmniGen~\cite{xiao2024omnigen} & 0.836 & 0.233 & - & 0.804 & - & - & - & - \\
PixWizard~\cite{lin2024pixwizard} & {0.845} & {0.248} & \textbf{0.069} & {0.798} 
& {0.884} & 0.265  & 0.063 & {0.876} \\
Explanatory Instructions~\cite{shen2024key} & 0.821 & {0.286} & 0.132 & 0.768 & 0.875 & {0.292} & 0.093 & 0.831 \\
UniReal~\cite{chen2024unireal} & 0.851 & 0.285 & 0.099 & 0.790 & 0.903 & \textbf{0.308} & 0.081 & 0.837 \\
\hline
\rowcolor{lightblue} \textbf{{\ours} (ours)} & \textbf{0.866} & \textbf{0.301} & 0.111 & \textbf{0.843}& \textbf{0.931} & {0.301} & 0.063 & \textbf{0.932} \\
\Xhline{1.2pt}
\end{tabular}
\end{table*}

To achieve robust and unified multi-modal image generation capabilities, CoLoGen is trained on large-scale synthetic datasets and a high-quality image-instruction-image triplet dataset through a multi-stage lifelong learning paradigm, as shown in Table \ref{tab:recipe}. In particular, the training set comprises our constructed synthetic datasets (3M), collected public datasets (22M), and high-quality in-house data (50K).

\noindent\textbf{Mask Inpainting.}
\label{sec:mask_inp}
In the endogenous pre-training stage, our goal is to learn rich visual concepts for individual ``objects'' and entire ``scenes''. To this end, we constructed a set of three million synthetic mask-inpainting data derived from JourneyDB \cite{sun2023journeydb}, incorporating three types of masks: random masks, object-shaped masks, and irregular object-shaped masks. Specifically, we use spacy \cite{spacy} to extract nouns from global captions and employ GroundingDino [19] and SAM [20] to segment target objects with corresponding masks, following the procedures in \cite{SeedDataEdit, xiao2024omnigen}. Additionally, we augment the training set with instance masks and global captions from the \cite{cocostuff} and \cite{zhou2017scene} datasets.

During training, we first generate random masks following \cite{suvorov2022resolution} and discard any masks whose intersection over union (IoU) with any object mask exceeds 0.3, thereby placing greater emphasis on scene-level concept learning. Next, inspired by \cite{yang2023paintbyexample}, we fit a Bessel curve to the bounding box of the masked object, uniformly sample 20 points along this curve, and connect them sequentially to form an irregular object-shaped mask which prevents instability in object generation caused by ill-defined mask shapes during testing. Finally, we sample random masks, object-shaped masks, and irregular object-shaped masks at a ratio of 20\%, 40\%, and 40\%, substantially enhance model’s robustness. 

\noindent\textbf{Image Grounding.}
\label{sec:img_grounding}
Image grounding \cite{shen2024key} involves identifying and highlighting specific object regions in an image based on textual prompts. The training data is sourced from RefCOCO \cite{refcoco}, RefCOCOg \cite{mao2016refcocog}, RefCOCO+ \cite{refcoco+}, and LVIS \cite{gupta2019lvis}. Following works \cite{lin2024pixwizard, shen2024key, zhao2024octopus}, we apply a variety of data augmentation strategies to construct three grounding tasks: (1) Box Detection Referring, where the target object is enclosed with bounding boxes of a specified color; (2) Mask Segmentation Referring, where the target object is covered with a specified color mask; and (3) Instance Detection Referring, where a random instance of the target class is enclosed with bounding boxes. 

\noindent\textbf{Controllable Image Generation.}
\label{sec:control_gen}
Following work \cite{xiao2024omnigen}, we collect the MultiGen (20M) \cite{qin2023unicontrol}, ADE20k \cite{zhou2017scene}, and COCOStuff \cite{cocostuff} datasets to support six visual condition controls, including Canny, Depth, HED, Lineart, and Segmentation. Notably, HED and Lineart are extracted online during training.

\noindent\textbf{Instruction Editing and Customized Generation.}
\label{sec:instruct}
Instruction-based editing and customized generation leverage simple textual interactions to efficiently modify or create images, offering substantial potential for practical applications. In this part, we consolidate multiple public editing datasets—MagicBrush (300K) \cite{zhang2024magicbrush} and OmniEdit (1.2M) \cite{OMNIEDIT} —along with the public customized-generation dataset Subject200K \cite{tan2024ominicontrol}. Furthermore, we incorporate 50K high-quality in-house editing data specifically curated to enhance the aesthetic appeal and realism of the generated images.

\section{Experiments}


\subsection{Training Recipe}

\begin{table*}[ht]
\centering
\renewcommand{\arraystretch}{1.0}
  \setlength\tabcolsep{2mm}
\caption{\textbf{Quantitative results for Controllable image generation} on MultiGen-20M, ADE-20K and COCOStuff. ``-'' indicates that the method does not report corresponding results. $\!\uparrow$ indicates that a higher value is better, while $\!\downarrow$ indicates that a lower value is better. 
} 
\label{tab:control}
\centering
\begin{tabular}{l|c|c|c|cc}
\Xhline{1.2pt}
\multirow{2}*{\textbf{Methods}} & \textbf{Canny-to-Image} &\textbf{Depth-to-Image} & \textbf{LineArt-to-Image}  & \multicolumn{2}{c}{\textbf{Seg-to-Image}} \\

 & CLIP-S↑ & RMSE↓ &SSIM↑  & \multicolumn{2}{c}{mIoU↑} \\
 & {MultiGen-20M} & {MultiGen-20M} & {MultiGen-20M} & COCO-Stuff & ADE20K \\
\hline
\rowcolor{gray!20}
\multicolumn{1}{l|}{\textsl{Specialist Models}}  &   &   & & & \\ 
ControlNet-SD1.5~\cite{controlnet}  &  {32.15} & 35.90 & - &27.46 &32.55 \\
T2I-Adapter-SD1.5~\cite{mou2024t2i}  &  31.71 & 48.40 & 63.94 & -&12.61 \\
Gligen ~\cite{li2023gligen} & - & 38.83 & - & - & 23.78 \\
Uni-ControlNet ~\cite{zhao2023uni} & - & 40.65 & - & - & 19.39 \\
UniControl ~\cite{qin2023unicontrol} & - & 39.18 & 70.54  & - & 25.44 \\
\hline
\rowcolor{gray!20}
\multicolumn{1}{l|}{\textsl{Generalist Models}}  &   &   & & & \\ 
OmniGen~\cite{xiao2024omnigen} & -  & \textbf{28.54} & - & - & \textbf{44.23} \\
PixWizard~\cite{lin2024pixwizard} & 32.01  & 33.83 & - & - & -  \\
Explanatory Instructions & 27.16  & 55.30 & - & - & - \\
\hline
\rowcolor{lightblue} \textbf{{\ours} (ours)}  & \textbf{33.31} & 31.79 & \textbf{77.96} & \textbf{29.43} & 42.82  \\
\Xhline{1.2pt}
\end{tabular}
\vspace{-1em}
\end{table*}

\begin{table}[ht]
\caption{%
    \textbf{Quantitative results for customized generation} on DreamBench~\cite{ruiz2023dreambooth}. 
}
\label{tab:comp_cus}
\vspace{-2mm}
\centering\scriptsize
\scalebox{1.2}
{
\begin{tabular}{lccc}
\Xhline{1.2pt}
\textbf{Model}   &  $\textbf{CLIP-T}\!\uparrow$   & $\textbf{CLIP-I}\!\uparrow$ &  $\textbf{DINO}\!\uparrow$   \\
\hline
\rowcolor{gray!20}
\multicolumn{1}{l}{\textsl{Specialist Models}}  & \multicolumn{3}{l}{} \\
Textual Inversion~\cite{mokady2023nulltextinversion}   & 0.255 & 0.780 &  0.569  \\
DreamBooth~\cite{ruiz2023dreambooth} & 0.305 & 0.803 &  0.668  \\
BLIP-Diffusion~\cite{li2024blipdiffusion} & 0.302 & 0.805 &  0.670  \\
ELITE~\cite{wei2023elite}  & 0.296  & 0.772 & 0.647  \\
Re-Imagen~\cite{chen2022reimagen} &  0.270  & 0.740 & 0.600  \\
BootPIG~\cite{purushwalkam2024bootpig} & 0.311 & 0.797 & 0.674  \\
%
\hline
\rowcolor{gray!20}
\multicolumn{1}{l}{\textsl{Generalist Models}}  & \multicolumn{3}{l}{} \\
OmniGen~\cite{xiao2024omnigen}  & 0.320  & 0.810 & 0.693 \\ 
UniReal\cite{chen2024unireal}   & \textbf{0.326}  & 0.806 & 0.702 \\
\hline
\rowcolor{lightblue} \textbf{{\ours} (ours)}   & 0.315 & \textbf{0.825} & \textbf{0.714} \\
\Xhline{1.2pt}
\end{tabular}
}
\end{table}

As illustrated in Fig. \ref{fig:intro}, the CoLoGen model training process consists of five steps: two steps of endogenous pre-training, one step of conditional injection learning and two steps of instruction-image alignment learning. The outline of all training datasets is shown in Table \ref{tab:recipe}.

\noindent\textbf{Concept–Localization Duality Pre-training. }
In the first stage of Concept–Localization Duality pre-training, we utilize three million synthetic data as mentioned in Sec \ref{sec:mask_inp} for the mask inpainting task. We apply the learning rate of 1e-4 for training 200K iterations with a global batch size of 256. We set the routing density $\rho$ of $expert_0$ equal to 1, and the loss of weight $\alpha_0 = 0$. Then, the second stage is trained on the task of image grounding. We apply the learning rate of 1e-4 for training 200K iterations with a global batch size of 256. The routing density $\rho$ of $expert_1$ is set to 0.8, and the loss weight is $\alpha_1 = 0.5$.

\noindent\textbf{Conditional Injection Learning. }
During this stage, the training dataset is composed of several publicly accessible datasets as mentioned in Sec \ref{sec:control_gen} builded on the task of controlable generation. The learning rate is 1e-4, the training iteration is set as 400K and the global batch size is 128. We set the routing density $\rho$ of $expert_2$ as 0.8, and the loss weight $\alpha_2$ as 0.5.

\noindent\textbf{Instruction-Image Alignment Learning. }In the final training stage, we first introduce Subject200k \cite{tan2024ominicontrol} for the training of customized image generation, and then fine-tune the model on the mixed dataset summarized in Table \ref{tab:recipe}. We train on the task of customized generation for 50K iterations with a global batch size of 128, followed by instruction editing for 200K iterations with the same batch size. For both tasks, we set the routing density $\rho$ of $expert_i$ to 0.8, and the loss weight $\alpha_i$ to 0.5.


%

\subsection{Main Results}
\noindent\textbf{Instruction Editing.}
We evaluate the CoLoGen on two Instruction editing benchmarks including the MagicBrush test split ~\citep{zhang2024magicbrush} and the Emu Edit test split ~\citep{sheynin2024emuedit}, which include diverse editing purposes, such as object addition, object removal, localized modifications, background alteration, etc. Following the evaluation metrics used both by Emu Edit and MagicBrush benchmarks, we evaluate four metrics: 1) $\text{CLIP}_{i}$: CLIP image similarity between source image and output image; 2) $\text{CLIP}_{out}$: CLIP text-image similarity between edited image and target caption; 3) $\ell_1$: $\ell_1$ distance between source image and output image; 4) DINO: DINO similarity between source image and output image. We compare our model against a variant of specialist and generalist models and report the quantitative results in Table \ref{tab:editing}. The findings indicate that CoLoGen achieves consistent and remarkable improvements across all both benchmarks in instruction adherence. The slightly higher L1 score compared to the SOTA model is expected, as substantial modifications between the input and output are anticipated under instruction-based editing. Moreover, L1 is not a particularly robust metric for this setting, consistent with works \cite{chen2024unireal, sheynin2024emuedit}.

\noindent\textbf{Controllable Image Generation.}
We use the dataset and script from \cite{li2024controlnet} to evaluate the ability on conditional control generation. For the {Segmentation-to-Image} condition, we report the $mIoU$ on COCO-Stuff and ADE20k benchmarks. As illustrated in Table \ref{tab:control}, we report CLIP similarity score CLIP-S for {Canny-to-Image} condition, SSIM metric for the {LineArt-to-Image} condition and RMSE for {Depth-to-Image} condition. CoLoGen achieves results comparable to those of SOTA controllable image generation methods.

\noindent\textbf{Customized Generation.}
Following OmniGen \cite{xiao2024omnigen}, we evaluate the single-entity customized generation capability using DreamBench \cite{ruiz2023dreambooth}, which comprises 750 prompts across 30 subjects. Similarly, we select only one input image per subject from the provided set of 4–7 images. Table \ref{tab:comp_cus} reports the DINO score, CLIP-I similarity, and CLIP-T similarity. Compared to specialist and generalist methods, CoLoGen achieves substantial improvements in DINO score and CLIP-I similarity, while obtaining comparable performance on CLIP-T similarity.

\begin{figure*}[ht]
\centering
\includegraphics[width=1.0\linewidth]{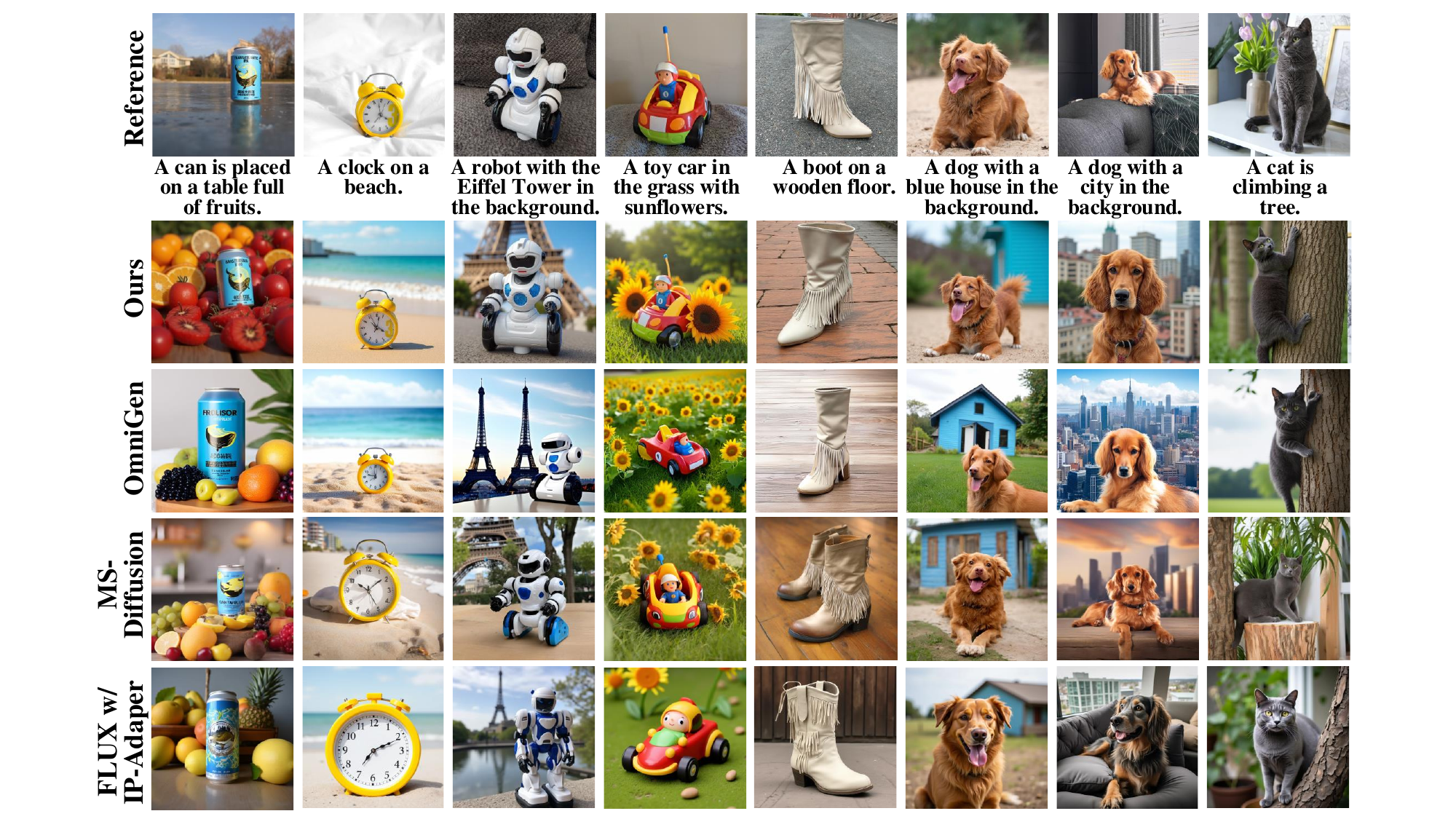}
\caption{Qualitative comparison for customized image generation. We compare with a series of public SOTA methods including OmniGen \cite{xiao2024omnigen}, MS-Diffusion \cite{wang2024ms}, and IP-Adapter \cite{Flux-IPadapter} on DreamBench \cite{ruiz2023dreambooth}. CoLoGen achieves remarkable performance even when trained on a limited amount of proprietary data, which can be attributed to the rich multimodal knowledge acquired by the model during the endogenous pre-training and conditional injection learning phases.}
\label{fig:customized}
\end{figure*}

\subsection{Ablation Studies}
\begin{table}[ht!]
\caption{%
    \textbf{Evaluation the contribution of the concept and localization representations on instruction-based editing (Magicbrush) and customized generation (Dreambench).} $\mathcal{R}_c$ denotes the concept representation and $\mathcal{R}_l$ denotes the localization representation. Metrics that exhibit a noticeable improvement over the baseline (w/o $\mathcal{R}_l$ \& w/o $\mathcal{R}_c$) are highlighted in \textcolor{syxblue}{\textbf{blue}}.
}
\label{tab:abl_representation}
\centering\scriptsize
\scalebox{1.2}{
\begin{tabular}{lccc}
\Xhline{1.2pt}
\textbf{Method}   &  $\textbf{CLIP-T}\!\uparrow$   & $\textbf{CLIP-I}\!\uparrow$ &  $\textbf{DINO}\!\uparrow$   \\
\hline
\rowcolor{gray!20}  Magicbrush &&& \\
w/o $\mathcal{R}_l$ \& w/o $\mathcal{R}_c$ & 0.260 & 0.889 & 0.901 \\
w $\mathcal{R}_l$               & 0.279 & \textcolor{syxblue}{\textbf{0.922}} & \textcolor{syxblue}{\textbf{0.927}}  \\
w $\mathcal{R}_c$               & \textcolor{syxblue}{\textbf{0.302}} & 0.881 & 0.905  \\
w $\mathcal{R}_c$ \& $\mathcal{R}_l$ (Co-training)      & 0.269 & \textcolor{syxblue}{\textbf{0.918}} & \textcolor{syxblue}{\textbf{0.922}} \\
w $\mathcal{R}_c$ \& $\mathcal{R}_l$ (CoLoGen)      & \textcolor{syxblue}{\textbf{0.301}} & \textcolor{syxblue}{\textbf{0.931}} & \textcolor{syxblue}{\textbf{0.932}} \\
\hline
\rowcolor{gray!20}  Dreambooth &&& \\
w/o $\mathcal{R}_l$ \& w/o $\mathcal{R}_c$ & 0.300 & 0.808 & 0.683 \\
w $\mathcal{R}_l$               & \textcolor{syxblue}{\textbf{0.310}} & \textcolor{syxblue}{\textbf{0.829}} & \textcolor{syxblue}{\textbf{0.707}}  \\
w $\mathcal{R}_c$               & 0.301 & 0.813 & \textcolor{syxblue}{\textbf{0.702}} \\
w $\mathcal{R}_c$ \& $\mathcal{R}_l$ (Co-training)     & \textcolor{syxblue}{\textbf{0.308}}  & 0.795 & 0.679 \\
w $\mathcal{R}_c$ \& $\mathcal{R}_l$ (CoLoGen)     & \textcolor{syxblue}{\textbf{0.315}}  & \textcolor{syxblue}{\textbf{0.825}} & \textcolor{syxblue}{\textbf{0.714}} \\
\Xhline{1.2pt}
\vspace{-7mm}
\end{tabular}
}
\end{table}

\begin{figure}[ht]
\centering
\includegraphics[width=1.0\linewidth]{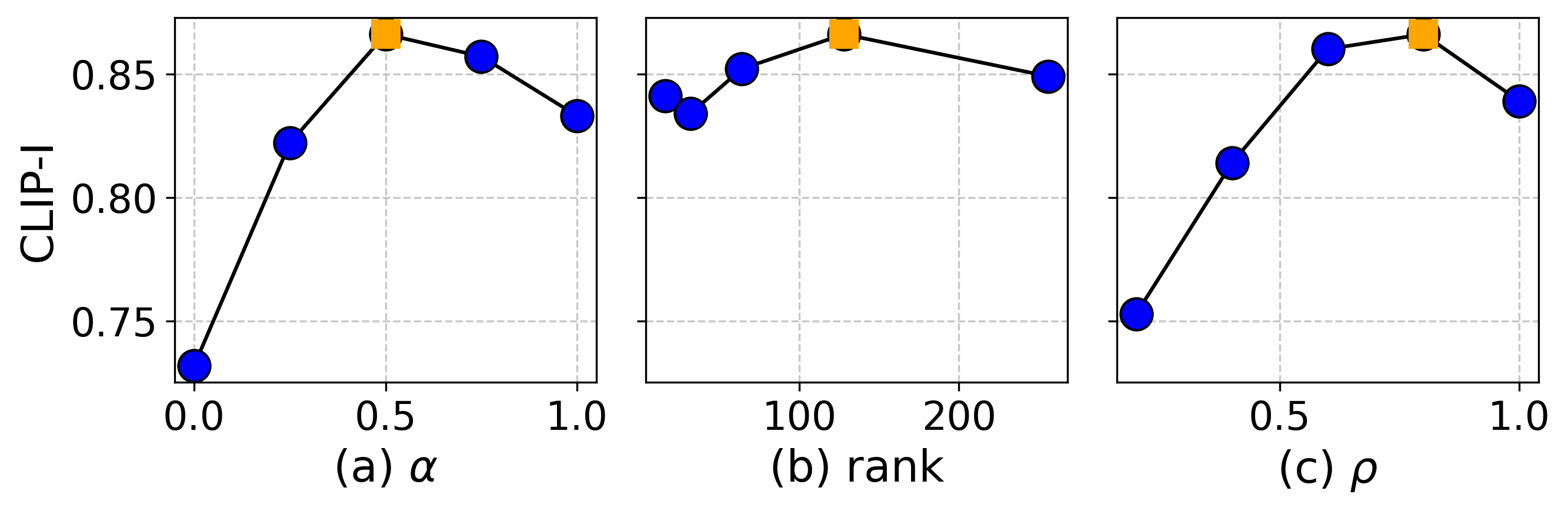}
\caption{Ablation studies for hyperparameter of lifelong strategy in the last stage, with final settings highlighted in \textcolor{orange}{orange}.
   (a) Impact of $\alpha$ on the weight for balancing the veteran gate routing supervision.
   (b) Influence of $rank$ on LoRA, where LoRA alpha weight defaults to twice the $rank$.
   (c) Impact of $\rho$, which denotes the rounting density of $expert_{N-1}$.}
\vspace{-7mm}
\label{fig:abl}
\end{figure}

\begin{figure*}[ht]
\centering
\includegraphics[width=1.0\linewidth]{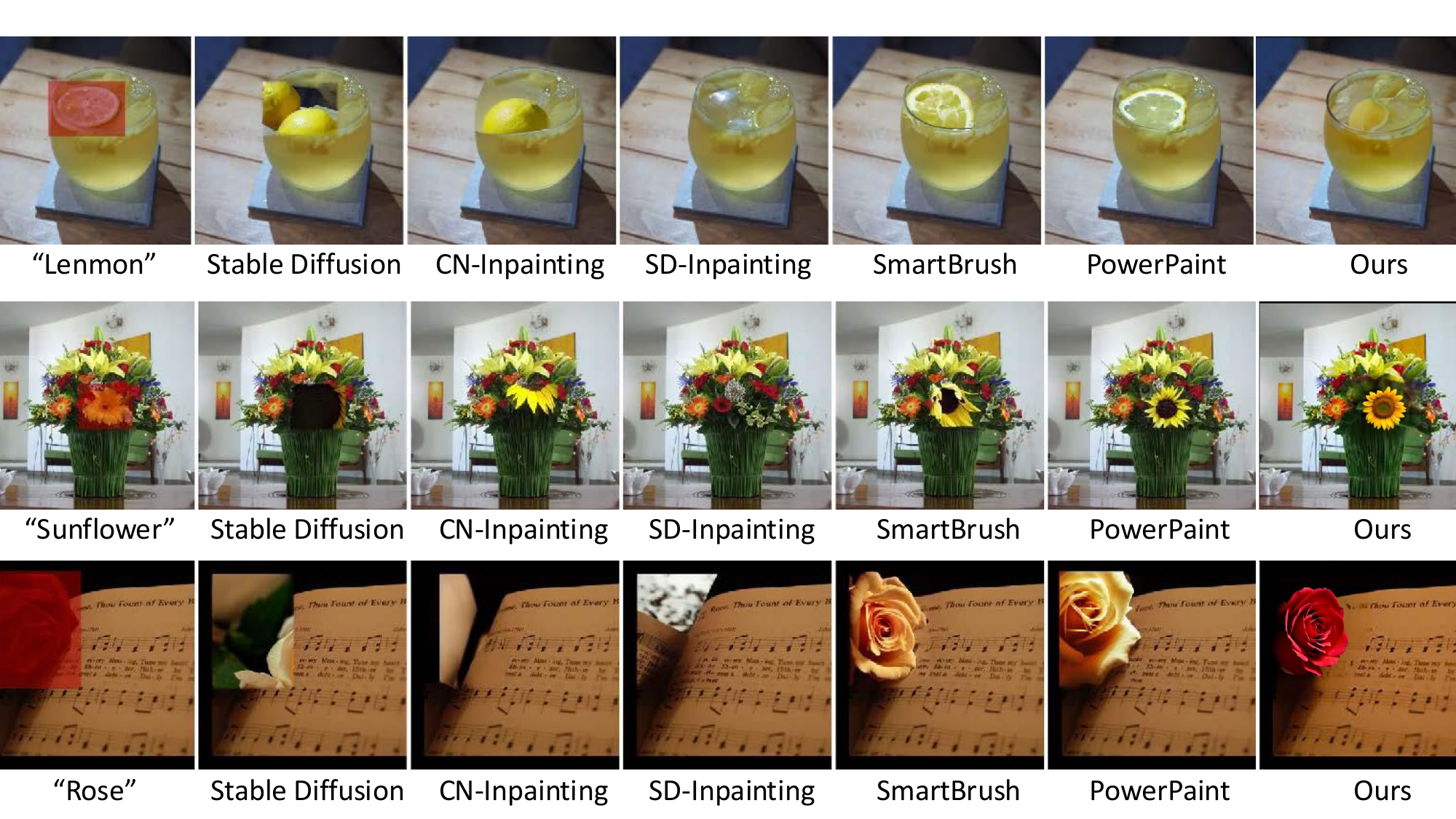}
\vspace{-2mm}
\caption{Qualitative comparisons with current state-of-the-art mask-inpainting methods. CoLoGen demonstrates robust text-following capabilities, and exhibits strong visual coherence between the mask area and the background.}
\vspace{-3mm}
\label{fig:mask_inp}
\end{figure*}

\noindent \textbf{The contribution of the concept and localization representations.}
We conduct a detailed ablation study on concept and localization representations across two high-level tasks: instruction-based editing and customized generation. As shown in Table \ref{tab:abl_representation}, our proposed CoLoGen model (with $\mathcal{R}_l$ and $\mathcal{R}_c$) consistently outperforms the baseline model (without $\mathcal{R}_l$ and $\mathcal{R}_c$) across six metrics on two benchmarks.

For the instruction-editing task (MagicBrush), incorporating $\mathcal{R}_c$ leads to a 0.042 improvement in CLIP-T, indicating that concept representations substantially enhance the model’s ability to follow instructions. Meanwhile, adding $\mathcal{R}_l$ improves both CLIP-I and DINO scores, demonstrating that localization representations significantly strengthen the model’s capability to identify and modify the intended regions while maintaining consistency in unedited areas.
Furthermore, in the customized generation task, the inclusion of localization representations also yields consistent improvements for CoLoGen over the baseline, highlighting general effectiveness across diverse generative objectives. 
On the other hand, multi-task co-training (with $\mathcal{R}_c$ \& $\mathcal{R}_l$), as illustrated in Figure \ref{fig:intro}(a), improves instruction following and fidelity only in one aspect, and even results in a decline in CLIP-I and DINO scores for customized generation compared with the baseline.

\noindent \textbf{Hyperparameters ablation.} We provide comprehensive ablation studies on hyperparameters, as illustrated in Fig. \ref{fig:abl}. All experiments are conducted during the instruction-editing fine-tuning stage, and we report the CLIP-I score for brevity. Fig. \ref{fig:abl}(a) demonstrates that veteran gate routing supervision effectively balances the utilization of specific experts and significantly improves CLIP-I performance. In Fig. \ref{fig:abl}(b), the LoRA $rank$ achieves optimal performance at 128. Fig. \ref{fig:abl}(c) indicates that $\rho$ performs best at 0.8, suggesting that the 20\% of experts trained during the early stage substantially contribute to downstream task performance.

\subsection{Qualitative Experiments}
\noindent\textbf{Customized Generation.} We present qualitative comparisons for customized image generation in Fig. \ref{fig:customized}. CoLoGen demonstrates superior performance compared with current state-of-the-art models in preserving details of the reference object, adhering to novel textual prompts, and maintaining consistency between the reference object and its background.

\noindent\textbf{Mask Inpainting.}
Mask inpainting enables the model to learn robust visual concept generation capabilities. Herein, we present qualitative comparisons with current state-of-the-art mask-inpainting methods. As illustrated in Fig. \ref{fig:mask_inp}, CoLoGen demonstrates strong inpainting capability which is trained on our large-scale synthetic datasets. Furthermore, the substantial endogenous capability developed during the pre-training stage equips CoLoGen with extensive multimodal knowledge for subsequent training stages.

\section{Limitation and Conclusion}
\noindent\textbf{Limitation.} While CoLoGen's Progressive Representation Weaving (PRW) architecture offers an efficient and adaptable design for various multi-modal to image generation models, it currently faces limitations concerning memory capacity as the number of tasks or the complexity of integrated experts scales up. 
Future work will focus on optimizing the PRW architecture for greater memory efficiency and scalability to handle an even broader range of challenging multi-modal tasks.

\noindent\textbf{Conclusion.} In this work, we introduce CoLoGen, a novel unified image generation framework designed to address the inherent concept–localization duality. 
By explicitly structuring tasks around conceptual and localization representations from our PRW model and employing a progressive staged training strategy, CoLoGen effectively mitigates representational conflicts and promotes cross-task complementarity. 
Extensive experiments across various benchmarks demonstrate that CoLoGen achieves competitive or superior performance compared to existing state-of-the-art methods. 
This work establishes a principled representational perspective for unified image generation, paving the way for more robust and versatile generative models in the future. 

\clearpage
{
    \small
    \bibliographystyle{ieeenat_fullname}
    \bibliography{main}
}

\clearpage
\setcounter{page}{1}
\maketitlesupplementary
\begin{appendices}

\begin{figure*}[t!]
\centering
\includegraphics[width=0.98\linewidth]{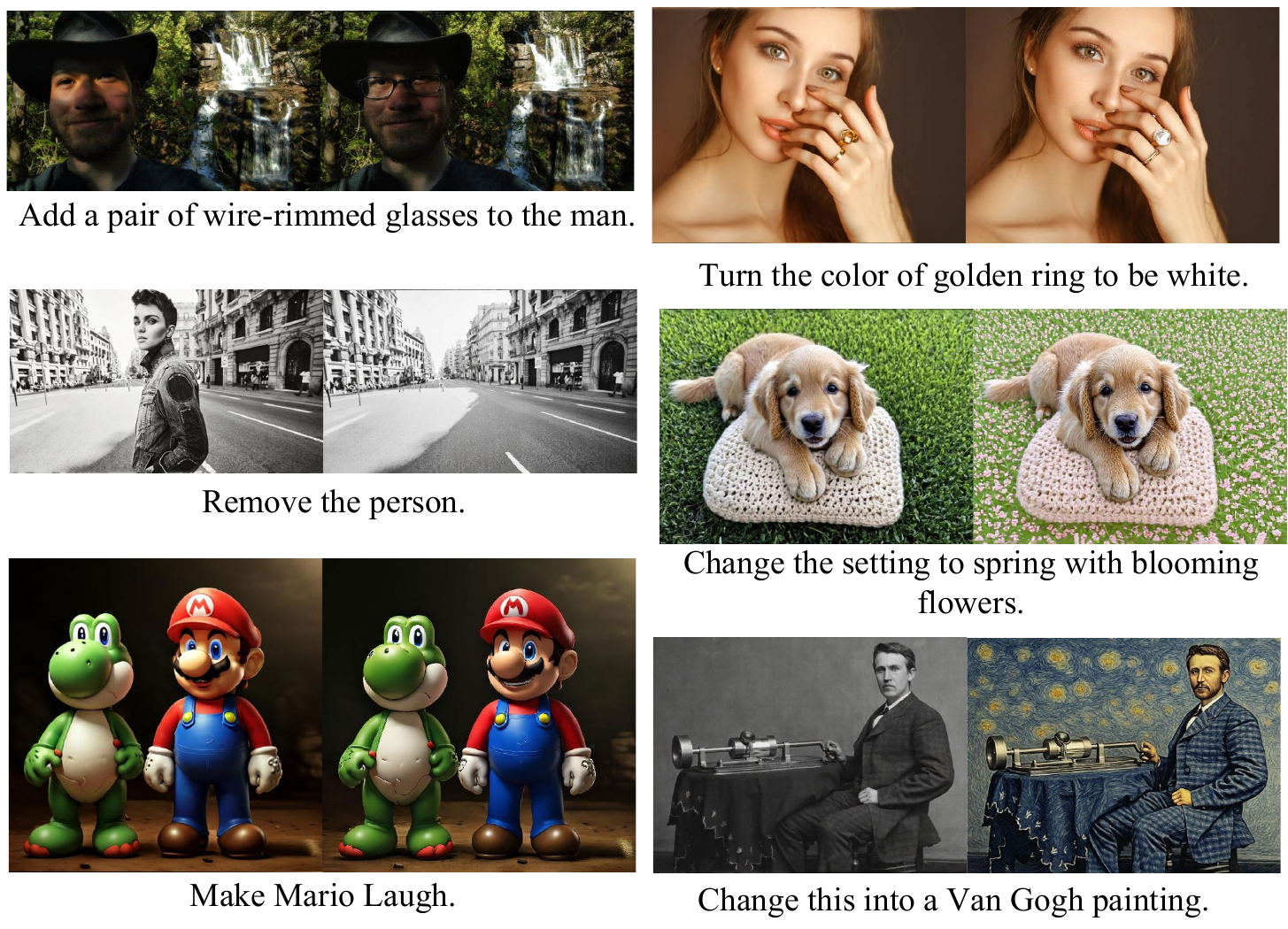}
\caption{Instructional editing results of our CoLoGen. Our method can adapt to various types of instructions, faithfully follow instructions while preserving the visual consistency of the input images, ensuring high-quality and coherent results.}
\label{fig:edit1}
\end{figure*}

\section{Related work}

\subsection{Unified Multi-modal to Image Generation}
Recent advancements in multi-modal image generation strive to consolidate diverse generation and editing tasks into unified frameworks, moving beyond task-specific pipelines. Early approaches often relied on distinct encoders or adapters for different conditions. For instance, \textbf{ControlNet}~\cite{zhang2023adding} and \textbf{T2I-Adapter}~\cite{mou2024t2i} introduced extensive external modules to guide pre-trained diffusion models. While effective, these methods face scalability issues when expanding to new tasks due to the linear growth of parameters.

To address this, recent works have focused on unified architectures. \textbf{Unified-IO 2}~\cite{lu2023unifiedio} and \textbf{Janus}~\cite{wu2024janus} demonstrate the power of autoregressive transformers in handling multi-modal inputs and outputs, though often at the cost of inference speed compared to diffusion models. In the diffusion domain, \textbf{OmniControl}~\cite{tan2024ominicontrol} and \textbf{DreamOmni}~\cite{xia2024dreamomni} integrate visual conditions directly into Diffusion Transformers (DiT), achieving spatial alignment with minimal parameter overhead. \textbf{OmniGen}~\cite{xiao2024omnigen} and \textbf{PixWizard}~\cite{lin2024pixwizard} further push the boundary by treating image generation and editing as a unified sequence generation problem, removing the reliance on external condition encoders entirely. Similarly, \textbf{UniReal}~\cite{chen2024unireal} treats image generation tasks as discontinuous video frames to capture real-world dynamics.

More recently, \textbf{Qwen-Image}~\cite{wu2025qwenimage} presents a large-scale diffusion foundation model emphasizing strong text rendering, multi-task training, and improved semantic–visual consistency for unified generation and editing.
\textbf{Query-Kontext}~\cite{song2025query} decouples multimodal reasoning from high-fidelity synthesis by leveraging a vision-language model to produce contextual query tokens that guide diffusion-based image generation and editing.
\textbf{Z-Image}~\cite{cai2025zimage} proposes an efficient single-stream diffusion transformer that unifies image generation and editing with scalable training, distillation, and accelerated inference.

However, these unified frameworks often struggle with what we identify as the \textit{Concept--Localization Duality}. Tasks like subject-driven generation require rich semantic concept encoding, whereas tasks like layout-to-image generation demand precise spatial structure. Naively training a single unified model often leads to representational conflict, where optimizing for semantic fidelity degrades spatial precision~\cite{chen2023octavius}. Unlike these approaches, \textbf{CoLoGen} explicitly decouples and progressively weaves these representations, ensuring high performance across both concept-heavy and localization-heavy tasks.

\subsection{Parameter-Efficient Composition and LoRA-MoE}
Low-Rank Adaptation (LoRA)~\cite{huang2023lorahub} has become the standard for parameter-efficient fine-tuning. To handle multi-task learning without catastrophic forgetting, recent research has explored Mixture-of-Experts (MoE) architectures combined with LoRA.

In the realm of Large Language Models (LLMs), \textbf{Octavius}~\cite{chen2023octavius} and \textbf{LoRAHub}~\cite{huang2023lorahub} propose routing mechanisms to dynamically select or compose LoRA modules for unseen tasks. In visual generation, \textbf{Mix-of-Show}~\cite{gu2023mixofshow} addresses the challenge of multi-concept personalization by fusing multiple LoRAs, while \textbf{ZipLoRA}~\cite{shah2023ziplora} attempts to merge content and style LoRAs by optimizing their orthogonality. \textbf{MoLE}~\cite{wu2024mole} applies a mixture of LoRA experts to select layer-wise adapters dynamically.
\textbf{ICEdit}~\cite{zhang2025icedit} enables instruction-based image editing via in-context generation, combining with LoRA-MoE.
While relevant, these methods typically employ static merging strategies or route based solely on input domains. They do not account for the evolving nature of representational needs during the diffusion process itself. \textbf{CoLoGen} advances this paradigm via our \textbf{Progressive Representation Weaving (PRW)}. Instead of static composition, we employ a time-step dependent "Veteran Gate" routing that dynamically balances expert usage. Crucially, our curriculum creates experts specialized specifically for \textit{Concept} versus \textit{Localization}, rather than just arbitrary data subsets, directly addressing the internal duality of generative tasks.

\section{More Results}

\subsection{Controllable Image Generation}

We expand our evaluation to recent state-of-the-art models built on stronger backbones (e.g., FLUX and SD3). While prior works typically report results under limited settings, we conduct a comprehensive comparison across both \textit{Canny} and \textit{Depth} conditions. As shown in Tab.~\ref{tab:backbone_comp}, our method consistently achieves the best overall performance across different metrics.

\begin{table}[t]
\centering
\scriptsize
\begin{tabular}{l|l|cc|cc}
\hline
& & \multicolumn{2}{c|}{Canny} & \multicolumn{2}{c}{Depth} \\
Method & Base & C-S$\uparrow$ & FID$\downarrow$ & SSIM$\uparrow$ & FID$\downarrow$ \\
\hline
UNIC-Adapter \cite{duan2025unicadapter} & SD3 & -- & 23.47 & 31.10 & -- \\
RealGen \cite{ding2024realgen} & CogV & -- & \textbf{17.50} & 35.0 & 23.40 \\
OmniControl \cite{xie2023omnicontrol} & FLUX & 30.60 & 20.63 & 39.0 & 27.26 \\
EasyControl \cite{zhang2025easycontrol} & FLUX & 28.60 & -- & 35.9 & 20.39 \\
\textbf{CoLoGen(Ours)} & FLUX & \textbf{33.31} & 18.20 & \textbf{40.1} & \textbf{19.56} \\
\hline
\end{tabular}
\caption{\textbf{Controllable image generation comparison} on recent backbone models.}
\label{tab:backbone_comp}
\vspace{-3mm}
\end{table}

\subsection{Customized Image Generation}

We further compare with recent large-scale customized generation methods, including Bagel and OmniGen2, on Subject-200k. Notably, these approaches are trained on substantially larger datasets (10M+ samples), whereas our method uses fewer than 1M samples. As reported in Tab.~\ref{tab:scale_comp}, CoLoGen achieves competitive or superior performance despite the significantly smaller training scale.

\begin{table}[t]
\centering
\scriptsize
\begin{tabular}{l|c|c|c|c}
\hline
Method & Data & DINO & C-I & C-T \\
\hline
OmniControl \cite{xie2023omnicontrol} & 200k & 0.684 & 0.799 & 0.312 \\
FLUX-IP-Adapter \cite{Flux-IPadapter} & 200k & 0.582 & 0.820 & 0.288 \\
\textbf{CoLoGen(Ours)} & 200k & \textbf{0.714} & 0.825 & \textbf{0.315} \\
UNO-FLUX \cite{wu2025unoflux} & 1M--5M & 0.760 & 0.835 & 0.308 \\
OmniGen2 \cite{wu2025omnigen2} & 10M+ & 0.749 & 0.830 & 0.314 \\
BAGEL \cite{de2006bagel} & 10M+ & 0.797 & \textbf{0.859} & 0.307 \\
\hline
\end{tabular}
\caption{\textbf{Customized image generation comparison} under different training data scales.}
\label{tab:scale_comp}
\vspace{-3mm}
\end{table}

\subsection{Image Editing Benchmark}

We additionally evaluate on the recent \textbf{GEdit-Bench} full set. As shown in Tab.~\ref{tab:GEdit-Bench}, CoLoGen achieves the best G\_SC score and remains competitive across other editing quality metrics, demonstrating strong generalization ability in image editing tasks.

\begin{table}[t]
\centering
\scriptsize
\begin{tabular}{l|c|c|c}
\hline
& \multicolumn{3}{c}{\textbf{GEdit-Bench (Full Set)} $\uparrow$} \\
Method & G\_SC & G\_PQ & G\_O \\
\hline
Step1X-Edit \cite{liu2025step1xedit} & 7.66 & 7.35 & 6.97 \\
BAGEL \cite{de2006bagel} & 7.36 & 6.83 & 6.52 \\
FLUX.1 Kontext \cite{labs2025fluxkontext} & 7.02 & 7.60 & 6.56 \\
Qwen-Image \cite{wu2025qwenimage} & 8.00 & \textbf{7.86} & \textbf{7.56} \\
\textbf{CoLoGen (Ours)} & \textbf{8.03} & 7.15 & 7.31 \\
\hline
\end{tabular}
\caption{\textbf{Results on GEdit-Bench (Full Set)}.}
\label{tab:GEdit-Bench}
\vspace{-10mm}
\end{table}

\begin{figure*}[htbp]
\centering
\includegraphics[width=0.98\linewidth]{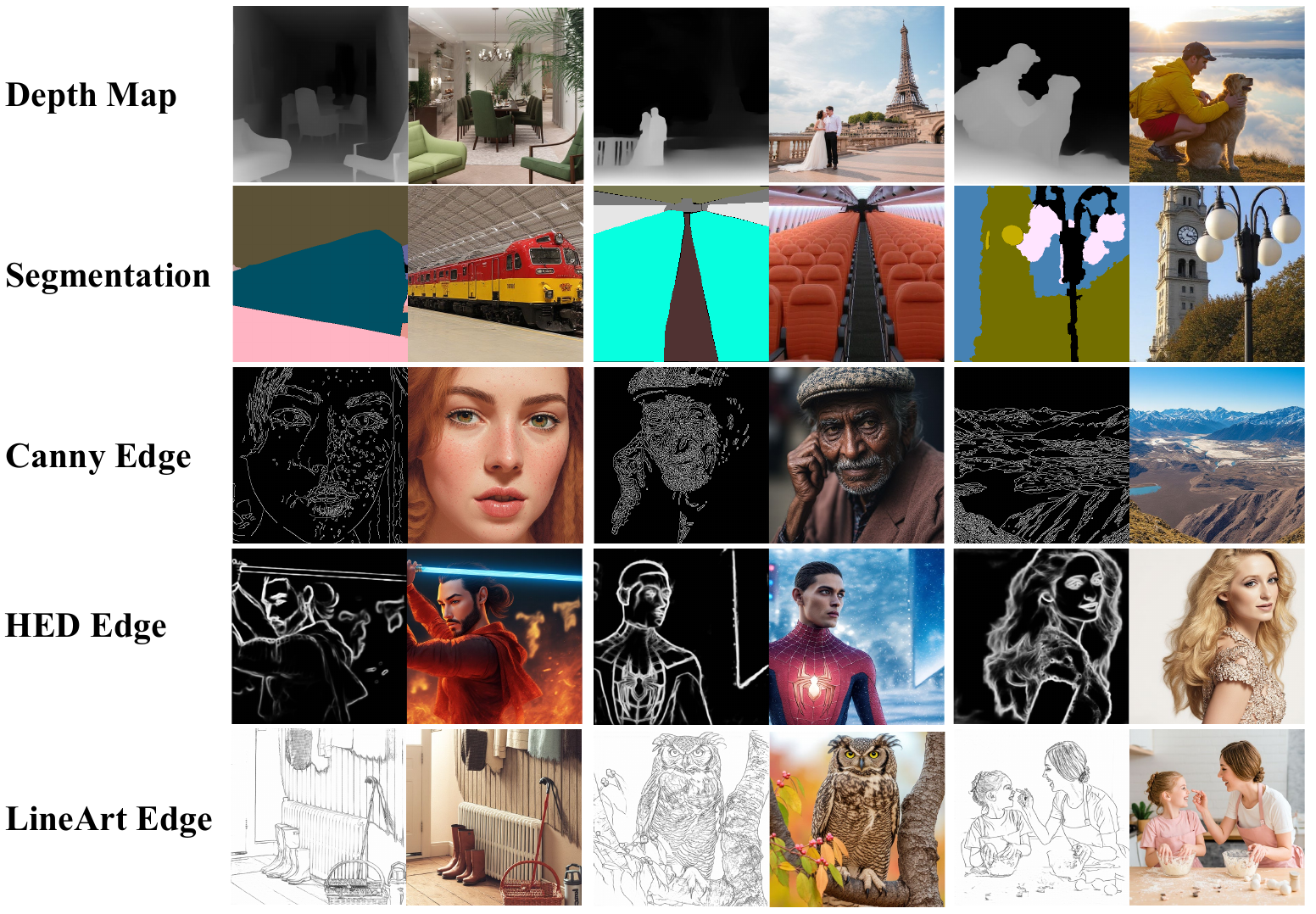}
\caption{Controllable generation results of our CoLoGen.}
\label{fig:edit2}
\end{figure*}

\begin{figure}[t]
\centering
\includegraphics[width=1.0\linewidth]{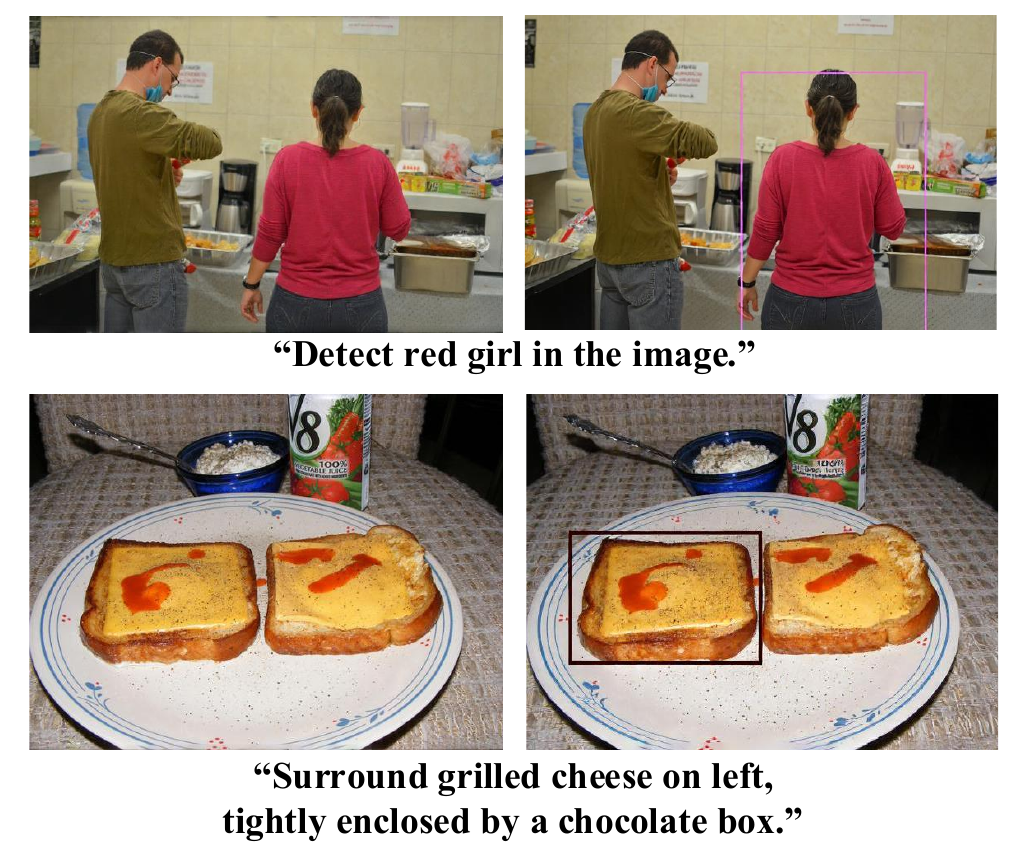}
\vspace{-4mm}
\caption{Visual examples from the image grounding task demonstrate that CoLoGen, after undergoing endogenous pre-training, exhibits highly accurate visual localization capabilities.}
\vspace{-4mm}
\label{fig:rec}
\end{figure}

\section{Visualization}
\subsection{Instruction Editing}
We provide expanded visual examples on the Instruction Editing benchmark in Figure~\ref{fig:edit1}. The results demonstrate CoLoGen's versatility in handling diverse editing instructions, ranging from localized object manipulation to global stylistic changes. 
These results validate that our Instruction-Image Alignment stage effectively fine-tunes the synergy between concept and localization representations.

\subsection{Controllable Image Generation}
Figure~\ref{fig:edit2} showcases CoLoGen's performance on the Controllable Image Generation benchmark under various spatial conditions, including Depth maps, Segmentation masks, Canny edges, HED edges, and LineArt.
The visualization highlights the effectiveness of the \textit{Localization Representation} ($R_l$) acquired during the endogenous pre-training.

\noindent\textbf{Image Grounding.}
CoLoGen acquires precise intent localization capabilities for the Image Grounding task during endogenous pre-training. The visualization in Fig. \ref{fig:rec} demonstrates that the model possesses robust object perception abilities and can accurately detect the referring instance, significantly enhancing its stability on complex tasks (e.g., instruction-based editing and customized generation).



\end{appendices}
\end{document}